\documentclass{bmvc2k}
\usepackage{amsfonts}

\title{Staged Contact-Aware Global Human Motion Forecasting}

\addauthor{Luca Scofano}{scofano@diag.uniroma1.it}{1}
\addauthor{Alessio Sampieri}{sampieri@diag.uniroma1.it}{1}
\addauthor{Elisabeth Schiele}{	elisabeth.schiele@tum.de }{2}
\addauthor{Edoardo De Matteis}{dematteis@di.uniroma1.it}{1}
\addauthor{Laura Leal-Taixé}{leal.taixe@tum.de}{2}
\addauthor{Fabio Galasso}{galasso@di.uniroma1.it}{1}

\addinstitution{
 Sapienza University of Rome
}
\addinstitution{
Technical University of Munich
}

\runninghead{Scofano et al.}{Staged Contact-Aware Global Human Motion Forecasting}

\def\etal{\emph{et al}\bmvaOneDot}

\newcommand{\model}{\textit{STAG}}

\begin{document}

\maketitle

\begin{abstract}

Scene-aware global human motion forecasting is critical for manifold applications, including virtual reality, robotics, and sports. The task combines human trajectory and pose forecasting within the provided scene context, which represents a significant challenge.

So far, only Mao \etal NeurIPS'22 have addressed scene-aware global motion, cascading the prediction of future scene contact points and the global motion estimation. They perform the latter as the end-to-end forecasting of future trajectories and poses. However, end-to-end contrasts with the coarse-to-fine nature of the task and it results in lower performance, as we demonstrate here empirically.

We propose a STAGed contact-aware global human motion forecasting (\textit{STAG}), a novel three-stage pipeline for predicting global human motion in a 3D environment. We first consider the scene and the respective human interaction as contact points. Secondly, we model the human trajectory forecasting within the scene, predicting the coarse motion of the human body as a whole. 
The third and last stage matches a plausible fine human joint motion to complement the trajectory considering the estimated contacts.

Compared to the state-of-the-art (SoA), \model{} achieves a 1.8\% and 16.2\% overall improvement in pose and trajectory prediction, respectively, on the scene-aware GTA-IM dataset. A comprehensive ablation study confirms the advantages of staged modeling over end-to-end approaches. Furthermore, we establish the significance of a newly proposed temporal counter called the "time-to-go", which tells how long it is before reaching scene contact and endpoints. Notably, \model{} showcases its ability to generalize to datasets lacking a scene and achieves a new state-of-the-art performance on CMU-Mocap, without leveraging any social cues. Our code is released at: \href{https://github.com/L-Scofano/STAG}{https://github.com/L-Scofano/STAG}.
\end{abstract}

\begin{figure}[!htp]
\centering
\includegraphics[width=1\textwidth]{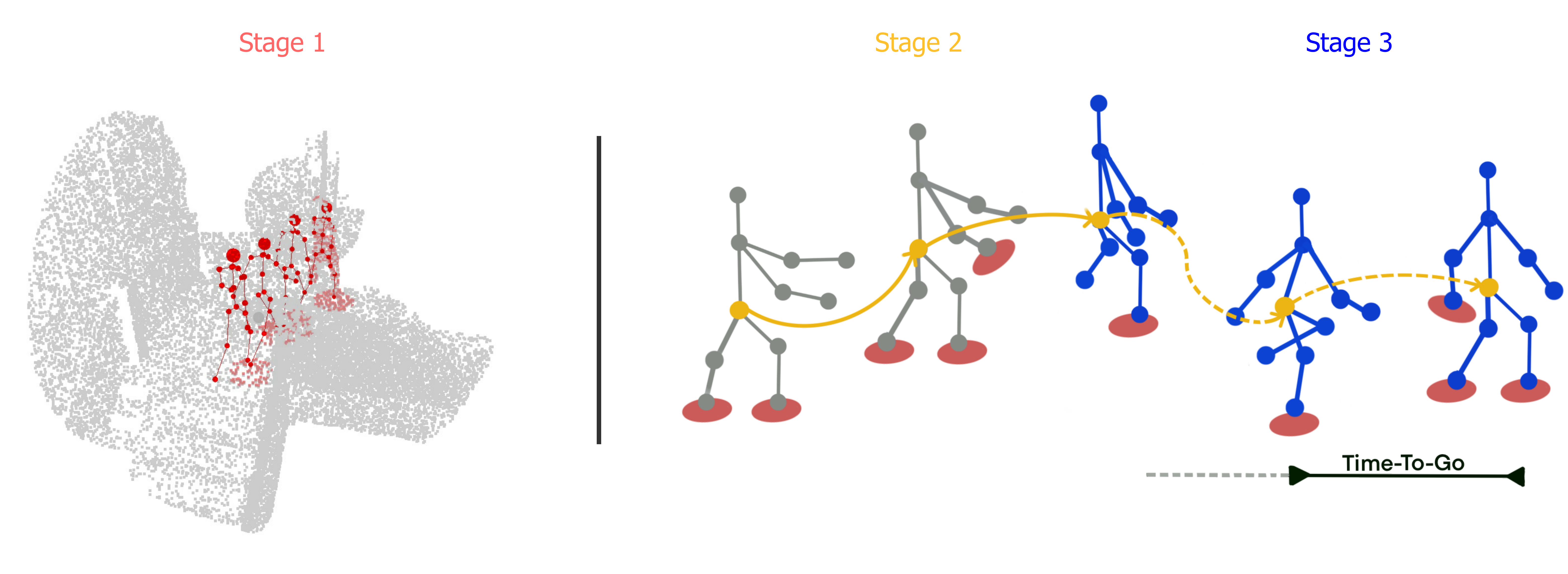}
\caption{
STAG forecasts scene-aware global human motion by three coarse-to-fine stages: (i) estimate the present and future contact points (light red) given the scene and the ground truth body joints (red); (ii) predict the future trajectory (dashed yellow), i.e.\ the future position of the root joints, given the past (solid yellow); (iii) predict the future body joints (blue) from the observed ones (gray).
Each stage of STAG conditions on the previous, so trajectory forecasting leverages future estimated contact points, and pose forecasting leverages both other estimates.
Awareness of the time-to-go (black arrow), the passing time between the current prediction and the end one, improves performance.}
\label{fig:teaser}
\end{figure}


\section{Introduction}

Humans are inherently predicting the near future at all times~\cite{Di_Luca2016-ro, Edwards2017-kl}.
As humans and machines coexist more, predicting human motion in the immediate future becomes critical for human-robot interaction, e.g., in industrial environments~\cite{Cho2017HumanGR, koppula13anticipating, sampieri2022pose} or breaking-in-time to avoid collisions~\cite{Paden2016ASO, Zhang2018HybridTP, Gonzlez2016ARO}.
Human motion forecasting generally includes \emph{local} pose forecasting~\cite{guo2022multi, guo2023back, mao2019learning, mao2020history, mao2021multi, sofianos21, yan2018spatial}, in which the joint locations are predicted with respect to the root joint, and \emph{global} pose forecasting~\cite{zhang2022wanderings, zhang2021we, mao2022contact}, which takes into account the positions of joints and the root in relation to a global coordinate system.

One common issue in human motion forecasting is the omission of the environment. It leads to contrived motion when the model is used in more realistic scenarios, such as \emph{ghost motions}, i.e., phasing through solid objects.
To our knowledge, \cite{mao2022contact} is the only work that accomplishes scene-aware global human forecasting.
They first process the scene and emphasize the human-scene interaction through contact points.
Subsequently, they employ an end-to-end approach to model individuals' trajectories and poses. Although contact points have shown effectiveness, employing end-to-end modeling for both trajectory and pose is suboptimal. The pose of an individual is influenced by their motion trajectory, interaction with the surrounding scene, and the pose in previous frames. However, the pose is typically not the underlying cause of the pathway. Essentially, global motion forecasting naturally aligns with a coarse-to-fine methodology that considers the scene, the trajectory, and the human pose.

We propose a novel model for \emph{STAGed contact-aware global human motion forecasting} (\model{}) that cascades three coarse-to-fine processing stages: (i) predicting the contact points, (ii) using them to forecast the trajectories, (iii) estimating the body pose (see Fig.\ref{fig:teaser}).
Our three-stage pipeline predicts the future motion autoregressively, conditioning each stage on the previous ones.
We condition the global motion on end goals and propose a time-to-go temporal encoding of the remaining duration until the endpoints are reached, i.e. informs the model on how many frames are missing.

We quantitatively evaluate the design choices of \model{} and compare them to the SoA on the available GTA-IM dataset~\cite{mao2022contact}.
Overall, we get up to a $21.1\%$ improvement on the path error with $16.2\%$ on average, while on the pose error, we get up to $5.4\%$ less error and $1.8\%$ on average.
We also show the generalizability of \model{} by testing it on CMU-Mocap~\cite{CMU}, a well-established multi-person dataset without scenes. To account for the missing scene, we only assume a planar ground. \model{} sets a new SoA without leveraging social cues, which SoA methods use~\cite{Wang2021MultiPerson3M, Adeli2020SociallyAC, giuliari2021transformer}. Overall, our contributions are threefold:
\begin{enumerate}
    \item We introduce a novel three-stage, coarse-to-fine model, which cascadedly processes the contact points, the trajectories, and the poses of people. 
    \item We introduce a learnable temporal counter for the time-to-go to align the predictions with the missing time before the endpoint.
    \item We perform a thorough analysis on GTA-IM~\cite{mao2022contact}, where we set a new SoA, and generalize \model{} to the \emph{scene-less} CMU-Mocap~\cite{CMU}. 
\end{enumerate}

\section{Related Work}

We discuss literature relating to the three core aspects of contact-aware global human motion forecasting: human-scene interaction (Sec. \ref{sec2:interaction}), trajectory forecasting (Sec. \ref{sec2:trajectory}), and human motion forecasting (Sec. \ref{sec2:hpf}).

\subsection{Human-Scene Interaction}\label{sec2:interaction}

Human motion forecasting is inherently influenced by the scene context in which it occurs, thus, considering the interaction between humans and their surroundings is crucial for motion forecasting. 

In motion synthesis, this shift towards including more contextual information can already be seen~\cite{zhang2022wanderings, wang2022towards, hassan2021stochastic, huang2023diffusion, chen2023unsupervised}, and some works in trajectory forecasting also consider contextual information~\cite{lee2022musevae, gartner2022trajectory, chen2022scept}. 
In human motion forecasting, scene information has been widely disregarded, with only a few works considering implicitly learning from the scene~\cite{corona2020context, cao2020long}.
However, this indirect modeling does not prevent \emph{ghost motion}, i.e. body parts passing through objects or the scene.
To the best of our knowledge, only \cite{mao2022contact} has investigated the explicit representation of human-scene interaction for human motion forecasting.
\cite{mao2022contact} proposes a two-stage pipeline,  first predicting future joint-scene distances, then using this information to predict the global pose.

Working with scene context requires data that enables the model to infer environmental clues.
3D point clouds provide dense information about surfaces and objects in the scene, which is ideal for human-scene interaction and trajectory forecasting.
\model{} elaborates on the idea of contact maps and adds a component of contextual knowledge through trajectory forecasting.

\subsection{Trajectory Forecasting}\label{sec2:trajectory}

Trajectory forecasting can be divided into two main categories: \emph{model-based} and \emph{model-free} approaches. 
Model-based approaches~\cite{berg2008recip, ellis2009pedestr, Helbing_1995, luo2008agent} impose physical constraints directly, while model-free approaches typically rely on implicitly learned physical plausibility~\cite{sadeghian2018sophie, ivanovic2019trajectron}. 
Some recent works~\cite{yue2022human, Antonucci_2022} have combined the two.
Model-free approaches employ a variety of deep learning techniques such as transformers~\cite{vaswani2017attention, yuan2021agentformer}, RNNs~\cite{Hewamalage_2021, mo2021graph} or GCNs~\cite{kipf2017semisupervised, mohamed2020social}. 
Deep learning approaches define the SoA on multiple benchmarks~\cite{yue2022human,mohamed2022social, liang2020simaug, liang2020garden,liu2021social}. \model ~follows best practices of SoA methods, adopting GCNs, attention, and trajectory endpoints.
Additionally, we are the first to propose a temporal encoding for the \emph{time-to-go}, informing the current model prediction of how long it is missing before the endpoint.

\subsection{Human Motion Forecasting}\label{sec2:hpf}
Human motion forecasting can be divided into (local) pose forecasting and global pose forecasting. 
Local pose forecasting~\cite{guo2022multi, guo2023back, mao2019learning, mao2020history, mao2021multi, sofianos21, yan2018spatial} only considers the position of the agent in relation to its root, while global pose forecasting~\cite{zhang2022wanderings, zhang2021we, mao2022contact} takes the absolute position within the given scene into account. 
Thus, global pose forecasting can be viewed as combining the trajectory and the (local) pose.
Many applications such as human-robot collaboration~\cite{Cho2017HumanGR, koppula13anticipating, sampieri2022pose}, autonomous driving~\cite{Paden2016ASO, Zhang2018HybridTP, Gonzlez2016ARO}, sports~\cite{Yuan2021AgentFormerAT, scofano2023latent}, augmented reality~\cite{starke2019neural} or animation~\cite{von2018recovering, ng2022learning} require knowledge about the global position of the agent in the scene.

Many human pose forecasting works use 2D image data only~\cite{becker2018red, zhan2019interaction, lerner2007crowds, ramanishka2018toward, rasouli2017they, ess2007depth}. 
However, in tasks such as industrial human-robot collaboration, where the agent's and objects' exact position in the scene is crucial, 3D data is often used~\cite{chang2019argoverse, caesar2020nuscenes, xue2019blvd}.
We consider a 3D point cloud for our task, as they offer a rich scene representation.
The agent in the scene can be represented as a graph of body joints or more complex representations such as meshes~\cite{SMPL-X:2019, hassan2019resolving, mahmood2019amass, guo2022generating}.

While delicate tasks may require a more specific human model, the skeletal representation (adopted in \model) suffices for human motion forecasting~\cite{guo2022multi, guo2023back, mao2019learning, mao2020history, mao2021multi, sofianos21, yan2018spatial}.

\section{Methodology}

\model{} is designed as a three-stage model, which we overview in Sec.~\ref{sec3:stag}. The modeling of each stage is detailed in Secs.~\ref{sec3:contact}-\ref{sec3:pose}.

\subsection{Proposed STAGed contact-aware global motion modelling}\label{sec3:stag}

Our novel approach for predicting contact-aware global human motion, named \model{}, is designed in a coarse-to-fine manner by using a three-stage pipeline (see Fig.~\ref{fig:model}). 
\model{} is composed of a first stage that computes the contact points between the 3D scene and the body (cf.\ Sec~\ref{sec3:contact}).
The second stage uses the information from the previous stage and the past root trajectory to predict its future trajectory (cf.\ Sec.~\ref{sec3:traj}). 
In the third stage,  the historical movement of the body, together with the contact points and human trajectory end goals, are used to predict the future global pose upon temporally encoding the time-to-go (cf.\ Sec~\ref{sec3:pose}).

\noindent\textbf{Notation.} We refer to $\mathbf{S}$ as the scene, $\mathbf{R}$ as the root joint trajectory, and to $\mathbf{M}$ as the global human motion. 
$\mathbf{S}$ represents a 3D scenes, where $\mathbf{S} \in \mathbb{R}^{N \times 3}$ contains $N$ points, each expressed as a triplet $(x,y,z)$.
$\mathbf{R}= [R_0, \dots, R_F]$ is a sequence of root trajectories, where $\mathbf{R}^T= [R_0, \dots, R_{T-1}]$ represents the observed ones and $\mathbf{R}^F= [R_T, \dots, R_{F}]$ the ones to predict. $R_i \in \mathbb{R}^{3}$ represents the 3D root coordinates.
Similarly $\mathbf{M}= [M_0, \dots, M_F]$ is a sequence of global body poses, where $\mathbf{M}^T= [M_0, \dots, M_{T-1}]$ represents the observed ones and $\mathbf{M}^F= [M_T, \dots, M_{F}]$ the ones to predict and $M_i \in \mathbb{M}^{V \times 3}$ represents the pose at timestamp $i$, consisting of $V$ joints expressed as 3D coordinates. 

\noindent\textbf{Staged processing.} In the first stage, the goal is to compute the contact points $\mathbf{C}$ defined as $[C_{0}, \dots, C_{T}, \dots, C_{F}]$, and $C_i \in \mathbb{R}^{V \times 4}$~\cite{mao2022contact} consisting of $V$ points expressed as 4D coordinates, triplet $(x,y,z)$ and one value $\{0,1\}$ to indicate whether it is a contact point or not (cf.\ Sec~\ref{sec3:contact}). The second stage predicts the future trajectory $\mathbf{R}^F$  given the historical root coordinates $\mathbf{R}^T$, and the contact points $\mathbf{C}$ (cf.\ Sec~\ref{sec3:traj}). In the third stage,  the objective is to predict the future body poses $\mathbf{M}^F$ by using $\mathbf{M}^T$, $\mathbf{C}$ and $\mathbf{R}^F$ (cf.\ Sec~\ref{sec3:pose}). 

\subsection{Contact Point Estimation} \label{sec3:contact}
\begin{figure}[!t]
\centering
\includegraphics[width=1\textwidth]{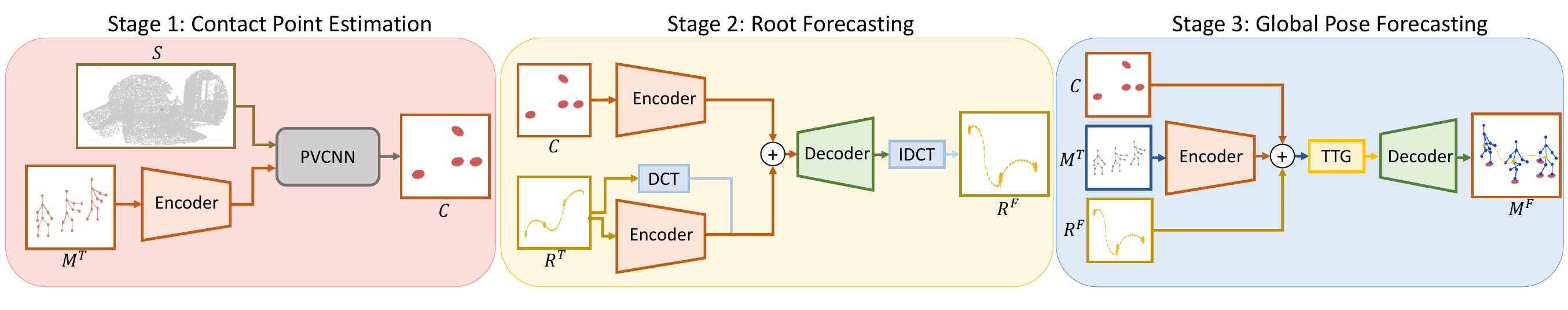}
\caption{Overview of \model{}'s three-staged pipeline.
Stage 1 takes the scene and the human motion in input and predicts future interactions as contact points. Stage 2 feeds them to a trajectory forecasting model for a coarse prediction, and Stage 3 then refines it to predict future human poses.}
\label{fig:model}
\end{figure}

In the first stage, the goal is to predict the contact points $\mathbf{C}$ between the global human body motion $\mathbf{M}$ and the scene $\mathbf{S}$ (See Fig.~\ref{fig:model}).
We use Point-Voxel CNN (PVCNN)~\cite{Liu2019PointVoxelCF} to model the scene $\mathbf{S}$ as a point cloud and encode $\mathbf{M}$ as a spatio-temporal graph~\cite{sofianos21}, to capture the movement's proprieties. 
Following~\cite{mao2022contact}, we first compute the distance matrix $\mathbf{D} \in \mathbb{R}^{TV \times N}$ where each term represents the Euclidean distance between each joint in time $TV$ and the $N$ points in the scene.
Since $\mathbf{D}$ is based on distances, it is smooth over time. 
We adopt a temporal encoding strategy of $\mathbf{D}$ based on the Discrete Cosine Transform (DCT)~\cite{mao2019learning}.

To leverage the DCT representation, we reformulate this problem by learning a mapping from the DCT coefficients of the past distance matrix $\mathbf{D}$ to those of the future one $\mathbf{\hat{D}}$. Following~\cite{mao2022contact}, we leverage  PVCNN~\cite{Liu2019PointVoxelCF} to encode the 3D scene $\mathbf{S}$, as well as the encoded motion $\mathbf{M}^T$ and $\mathbf{D}$'s DCT coefficients. 

Following leading pose forecasting literature, we use Graph Convolutional Networks~\cite{kipf2017semisupervised} to encode the motion. Similarly to previous works~\cite{sofianos21, xu22stars}, we define a \emph{spatial adjacency matrix} as $A_s \in \mathbb{R}^{T \times V \times V}$ to model the connections between joints and a \emph{temporal adjacency matrix} $A_t \in \mathbb{R}^{V \times T \times T}$ to capture the temporal relationships. 

\begin{equation}
    \bar{\mathbf{M}}^T = \sigma(A_s A_t \mathbf{M}^T W)
    \label{eq:gst}
\end{equation}

We aim to obtain a latent vector representing the entire movement sequence and serving as a conditioning variable. To compress spatial and temporal information, we propose using two separate MLPs, $MLP_S$, and $MLP_T$:
\begin{equation}
    \Tilde{\mathbf{M}}^T = MLP_S(MLP_T( \bar{\mathbf{M}}^T))
    \label{eq:gcnmlp}
\end{equation}
Each MLP consists of two linear layers and an equal number of activation functions. From now on, we will refer to this encoding technique as \textit{GCN-MLP}.
In summary:
\begin{equation}
    \hat{\mathbf{D}} = IDCT(\mathbf{M}^T+\textit{f}(\mathbf{S},DCT(\mathbf{D}),\Tilde{\mathbf{M}}^T))
\end{equation}
Where \textit{f} represents the trainable point-cloud encoder~\cite{Liu2019PointVoxelCF}.
Ultimately, we reconvert the distance matrix $\mathbf{\hat{D}}$ to the contact points as in~\cite{mao2022contact} resulting in the predicted contact points $\mathbf{C}$.

\subsection{Root Forecasting} \label{sec3:traj}

In the second stage, we propose to predict the person's trajectory to account for future global motion. We achieve it by predicting the future root joint $\mathbf{R}^F$ from the past $\mathbf{R}^T$. The second stage integrates the scene contacts $\mathbf{C}$, estimated in stage one.
$\mathbf{R}^T$ is encoded twice, once by using DCT and secondly by using the encoder described in Sec.~\ref{sec3:contact}. The formulation is similar, however, $\mathbf{M}^T$ gets changed with $\mathbf{R}^T$ in Eq.~\eqref{eq:gcnmlp} and the number of nodes $\textit{V}=1$, resulting in $\Tilde{\mathbf{R}}^T$. The same encoding technique is used for the contact points, where $\mathbf{M}^T$ gets changed with $\mathbf{C}$ in Eq.~\eqref{eq:gcnmlp} and results in $\Tilde{\mathbf{C}}$. The latter encodings are concatenated and fed to an MLP, which decodes the feature dimension $C$ and outputs $\hat{\mathbf{R}} \in \mathbb{R}^{T \times V \times C}$. 
\begin{equation}
    \hat{\mathbf{R}}^T = MLP(DCT(\mathbf{R}^T) \mathbin\Vert \Tilde{\mathbf{R}}^T \mathbin\Vert \Tilde{\mathbf{C}})),
\end{equation}
where $\mathbin\Vert$ indicates a concatenation operation.
Lastly, the $IDCT$ reverts the transformation process to trajectories so that $\mathbf{R}^F = IDCT(\hat{\mathbf{R}}^T)$. 

\subsection{Global Pose Forecasting}\label{sec3:pose}

For the third stage, we utilize the forecasted root trajectory $\mathbf{R}^F$ and the contact points $\mathbf{C}$ obtained from the preceding stages (see Fig.~\ref{fig:model}) as inputs. 
It enables us to predict the future pose and refine the trajectory, ultimately yielding the future global motion of the agent. We also encode the past body motion $\mathbf{M}^T$ as in Sec. \ref{sec3:contact} and concatenate the latter information.
The decoding occurs autoregressively, where each future timeframe $\{i\}_{i=T}^{F}$ of the predicted body motion $\mathbf{M}_{i}^{F}$ is computed sequentially.
We propose to temporally encode the scene contact points and the trajectory endpoints to raise the model understanding of the time-to-go, i.e.\ how long before it reaches them.
At each $i$, we also concatenate the root's position $\mathbf{R}_{e}^{F}$, and the contact points $\mathbf{C}_{e}^{F}$ at the last frame $F$ as end goal conditioning variables. Where respectively, $\mathbf{R}_{e}^{F} \in \mathbb{R}^{C}$ and $\mathbf{C}_{e}^{F} \in \mathbb{R}^{V \times C}$.
\begin{equation}
    {HM}_{i}^{F} = MLP(\Tilde{M}^{T} \mathbin\Vert {R}_{i+1}^{F} \mathbin\Vert \mathbf{R}_{e}^{F} \mathbin\Vert \Tilde{C}_{i+1} \mathbin\Vert \Tilde{\mathbf{C}}_{e}^{F}),
\end{equation}
where ${HM}_{i}^{F}$ is the the embedding at time $\{i\}_{i=T}^{F}$. Then we add the TE and decode the global body pose.

\paragraph{Time-to-go Temporal Encoding}
 To insert time context, we use a learnable temporal encoder $\mathcal{T}_e$ to encode the time-to-go and add it to ${HM}_{i}^{F}$.
 During the autoregressive process, $\mathcal{T}_e$ measures how long is missing before the contact and endpoints are reached. 
 To decode the global motion we use an MLP layer. In summary:

\begin{equation}
 M^i = MLP({HM}^{i}+\mathcal{T}_e).
\end{equation}

\section{Experiments}
In this section, we detail the dataset and metrics, compare it to the current SoA~\cite{mao2022contact}, and perform an extensive ablation on the staged modeling design and its components. Furthermore, we show how our model compares with SoA.

\paragraph{Dataset.}
The GTA-IM dataset~\cite{cao2020long} is a large-scale synthetic dataset that captures human-scene interactions, which consists of 50 different characters performing various activities in 7 scenes. 
We use~\cite{mao2022contact}'s proposed pre-processing, employing 4 of the scenes as our training set, the remaining 3 as the test set, and 21 out of the 98 human joints provided by the dataset. 
Videos are recorded at 30fps, and we train our models to observe the past 30 frames and predict the future 60.
We evaluate \model{} in all its stages and outperform~\cite{mao2022contact}.

\cite{mao2022contact} also considers PROX~\cite{hassan2019resolving} but they do not distribute the pre-processed scene-to-pointcloud nor the code for pre-processing. PROX is a real dataset captured using a Kinect-One sensor, and it contains noise at frames (e.g.\ jittering and corrupted pixels) and in time (missing frames). Upon best efforts, we could not replicate the pre-processed pointcloud, so we could not use it for comparison.

\paragraph{Metrics.}
The first stage is evaluated by the $L_2$-norm between our predicted contact point and the ground truth.
For the second and third stages, we consider the \emph{Mean Per Joint Positional Error} (MPJPE) across all joints and all the future timeframes~\cite{mao2022contact}. The global movement is called Path Error, and the Pose Error represents the local body movement.

\paragraph{State-of-the-art models and selected baserows.} We evaluate \model{} on the GTA-IM dataset and compare it with the current leading techniques. LTD~\cite{mao2019learning} utilizes a graph convolutional network to encode motion representations in frequencies. DMGNN~\cite{li2020dynamic} employs a \textit{dynamic multiscale} GNN for sequence encoding, with a decoder based on GRU. SLT~\cite{wang2021synthesizing} focuses on motion synthesis and employs an autoencoder architecture consisting of a multilayer perceptron as the initial stage, followed by motion generation using LSTM. The top-performing technique is Mao \etal~\cite{mao2022contact}, which combines MLP and RNN for motion encoding and employs an iterative prediction approach.

\subsection{Comparison against SoA}

\paragraph{First Stage - Contact points estimation.}
Table \ref{tab: s1} is not present in \cite{mao2022contact}; thus, we ran their first stage and compared it to ours (Sec.~\ref{sec3:contact}). 
We have an overall 9.2\% improvement, and it is due to our body movement's encoder, which more accurately extracts the latent representations.

\begin{table}[!htp]
\centering
\scriptsize
\begin{tabular}{l|rrrrrr} \hline
{} & \multicolumn{5}{c}{$L_2$-norm (mm)} \\
&0.5s &1s &1.5s &2s &mean \\
\hline
Mao \etal~\cite{mao2022contact} &26.2 &45.5 &67.5 &96 &47.8 \\
\model &\textbf{24.3} &\textbf{41.9} &\textbf{61.6} &\textbf{86,2} &\textbf{43.6} \\ \hline
\end{tabular}
\caption{Distance between the predicted contact points and the ground truth ones.}
\label{tab: s1}
\end{table}
\paragraph{Second Stage - Global pose error.}
This section focuses on our second stage's impact on the Path Error.
As in \cite{mao2022contact}, we experiment with three configurations of our model: (i) no contact point to condition on, (ii) conditioning on the predicted contact points, and (iii) on the ground truth ones. With the original configuration (i), we note a decline in performance, indicating that the second stage necessitates supplementary contextual information for accurate operation. This emphasizes the importance of considering the scene when predicting overall bodily motion.
When the predicted contact points (ii) are added, \model{} has a 16.2\% more accurate prediction over the path error. Such improvement increases when considering the GT contact points (iii), reaching a 21\% decrease in mean over path error. (iii) also highlights that having precise contact points coming from stage one can significantly improve the overall performance of the second stage.

\paragraph{Third Stage - Local pose error.}
As in the previous paragraph, we consider: (i) no contact point to condition on, (ii) conditioning on the predicted contact points, and (iii) on the ground truth ones.
In this case, we outperform \cite{mao2022contact} in all settings, reaching 1.8\% improvement in (iii) and 4\% when considering GT contact points. It demonstrates how our body movement encoder is more capable of creating reasonable latent representations. While the improvement in pose may not be as pronounced as the improvement in path, it is crucial to consider the 3D nature of the scenario and ensure coherent body movements by accounting for the surroundings. With \model{}, we observe an overall enhancement in both path and pose compared to the SoA methods. The staged pipeline assigns equal importance to both tasks, leading to these improvements.

\begin{table}[!htp]\centering

\scriptsize
\begin{tabular}{l|ccccc|ccccc}
\hline
\textbf{} &\multicolumn{5}{c}{\textbf{Path Error (mm)}} &\multicolumn{5}{|c}{\textbf{Pose Error (mm)}} \\
\textbf{Models} &\textbf{0.5s} &\textbf{1s} &\textbf{1.5s} &\textbf{2s} &\textbf{mean} &\textbf{0.5s} &\textbf{1s} &\textbf{1.5s} &\textbf{2s} &\textbf{mean} \\
\hline
 LTD~\cite{mao2019learning} & 67.0 & 119.3 & 207.6 & 375.6 & 147.4 & 67.5 & 93.8 & 98.9 & 103.5 & 80.5 \\
DMGNN~\cite{li2020dynamic} & 82.7 & 158.0 & 227.8 & 286.9 & 156.2 & 47.5 & 69.1 & 85.6 & 95.3 & 64.9 \\
SLT$^*$~\cite{wang2021synthesizing} & \textbf{45.9} & 117.0 & 186.7 & 267.1 & 121.8 & 70.8 & 181.4 & 150.2 & 196.0 & 112.6 \\
\hline
Mao \etal~\cite{mao2022contact} \textit{w/o contact} & 61.1 & 111.7 & 171.0 & 249.0 & 118.8 & 57.8 & 74.8 & 82.4 & 98.1 & 68.2 \\ 
Mao \etal~\cite{mao2022contact} \textit{w/ pred contact} & 58.0 & 103.2 & 154.9 & 221.7 & 108.4 & 50.8 & 67.5 & 75.5 & \textbf{86.9} & 61.4 \\
\textcolor{gray}{Mao \etal~\cite{mao2022contact} \textit{w/ GT contact}} & \textcolor{gray}{52.4} & \textcolor{gray}{77.8} & \textcolor{gray}{95.8} & \textcolor{gray}{129.5} & \textcolor{gray}{74.1} & \textcolor{gray}{49.8} & \textcolor{gray}{64.8} & \textcolor{gray}{70.4} & \textcolor{gray}{78.3} & \textcolor{gray}{58.2} \\
\hline
\model{} \textit{w/o contact} & 64.0 & 133.0 & 210.4 &   302.0 & 141 & 55.8 &    72.9 &    82.8 &    96.2 &    67.1 \\
\model{} \textit{w/ pred contact} & 55.4 & \textbf{89,6} & \textbf{127.9} & \textbf{179.3} & \textbf{92.3} & \textbf{48.1} & \textbf{65.3} & \textbf{75.6} & 88.2 & \textbf{60.3}\\
\textcolor{gray}{\model{} \textit{w/ GT contact}} & \textcolor{gray}{50.3} & \textcolor{gray}{65.1} & \textcolor{gray}{70.1} & \textcolor{gray}{99.2} & \textcolor{gray}{60.0} & \textcolor{gray}{46.9} & \textcolor{gray}{61.5} & \textcolor{gray}{68.0} & \textcolor{gray}{76.3} & \textcolor{gray}{55.6} \\ \hline
\end{tabular}
\caption{
Path and pose error on the output obtained by pipelining the second and third stages on GTA-IM dataset.}
\label{tab: main}
\end{table}

\subsection{Ablation study}

We perform ablative studies to explore our model's components extensively.
The results in Table \ref{tab: s2_ablation} consider GT contact points and refer to the metrics used in Table \ref{tab: main}.
\textbf{stages} indicates the training mode of the second stage module:
\emph{2-stage e2e} means that stages two and three are learned in an e2e fashion, as is done in \cite{mao2022contact}; \emph{2-stage ft.} indicates that stage two is pre-trained and fine-tuned during the training of stage thee; \emph{3-stage (\model{})} is our proposed pipeline. \textbf{end} indicates whether the endpoint is used in the third stage.
\textbf{TTG} is flagged if the proposed \emph{time-to-go} is used in the third stage. Regarding how many joints are used to compute the contact points, we conducted a dedicated ablation study outlined in Table 3. The \textbf{cont.} column indicates which joints we consider for contact. With ”all”, every body part is considered to estimate contact with the scene. With ”feet”, only the feet can generate contact points, while with ”feet, wrist”, we consider contact points involving both the feet and hands.

\begin{table}[!bp]
\centering
\scriptsize
\resizebox{\textwidth}{!}{
\begin{tabular}{r|cccc|ccccc|ccccc}
\hline
~ & \multicolumn{4}{c}{~} & \multicolumn{5}{|c|}{\textbf{Path Error (mm)}} &\multicolumn{5}{c}{\textbf{Pose Error (mm)}} \\
~ & \textbf{stages} & \textbf{end.} & \textbf{TTG} & \textbf{cont.} & \textbf{0.5s} &\textbf{1s} &\textbf{1.5s} &\textbf{2s} &\textbf{mean} &\textbf{0.5s} &\textbf{1s} &\textbf{1.5s} &\textbf{2s} &\textbf{mean} \\
\hline

1 & 2-stage e2e & $\times$ & $\times$ & all &55.8 &77.7 &87.5 &121.5 &71.3 &48.8 &64.15 &70.7 &77.9 &57.8 \\
2 & 2-stage ft. & $\times$ & $\times$ & all &53.6 &72.5 &83.7 &115.8 &68.4 &48.8 &64.2 &70.8 &77.9 &57.8 \\
3 & 3-stage (\model) & $\times$ & $\times$ & all &53.4 &72.4 &84.1 &117.8 &68.5 &48.8 &64.2 &70.9 &78 &57.8 \\
\hline

4 & 2-stage e2e & \checkmark & $\times$ & all & 55.7 &79.2 &95.2 &128.3 &75.3 &47.1 &61.8 &68.5 &76.7 &56 \\
5 & 2-stage ft. & \checkmark & $\times$ & all & 51.9 &68.7 &78.1 &113.1 &65 &47.1 &61.8 &68.5 &76.6 &55.9 \\
6 & 3-stage (\model) & \checkmark & $\times$ & all & 51.6 &68.3 &76.8 &108.4 &64 &47.1 &61.9 &68.4 &76.8 &56 \\
\hline

7 & 2-stage e2e & \checkmark & \checkmark & all & 53.8 &74.6 &87.2 &122.4 &70 &47.2 &61.9 &68.5 &76.9 &56 \\
8 & 2-stage ft. & \checkmark & \checkmark & all &50.8 &66.1 &72.6 &104.7 &61.6 &47.1 &61.8 &68.4 &76.6 &55.9 \\
9 & 3-stage (\model) & \checkmark & \checkmark & all & \textbf{50.3} &\textbf{65.1} &\textbf{70.1} &\textbf{99.2} &\textbf{60} &\textbf{46.9} &\textbf{61.5} &\textbf{68} &\textbf{76.3} &\textbf{55.6}  \\
\hline

10 & 3-stage (\model) & \checkmark & \checkmark & feet & 55.1 &79.9 &96.7 &136.5 &76.2 &47.5 &62.1 &68.7 &77.3 &56.3 \\
11 & 2-stage e2e & \checkmark & \checkmark & feet, wrist & 56.0 &81.4 &99.0 &135.7 &77.7 &47.1 &61.8 &68.4 &77.1 &56 \\
12 & 2-stage ft. & \checkmark & \checkmark & feet, wrist & 55.6 &82.0 &102.0 &140.0 &79 &47.1 &61.4 &68.9 &79.9 &56 \\
13 & 3-stage (\model) & \checkmark & \checkmark & feet, wrist & 56.9 &84.4 &104.9 &143.0 &81.0 &46.2 &61.4 &68.3 &77.1 &55.9 \\ \hline
\end{tabular}
}
\caption{Ablation study on the staged modeling.}
\label{tab: s2_ablation}
\end{table}

The performance comparison in rows 1-3 reveals that even without the inclusion of end goals or TTG, the three-stage pipeline surpasses the performance of the two-stage pipeline. When end goals are introduced (rows 4-6), the performance gap becomes more apparent as they contribute to improved global performance. Lastly, using TTG in autoregressive prediction introduces time context and significantly enhances the results (rows 7-9).
Moreover, it is preferable to consider the entire skeleton when calculating contact points, as the 3D scene is complex and involves multiple joints. Merely focusing on ground contact points (e.g., feet) or the most probable contact points (e.g., feet and hands) leads to unsatisfactory outcomes, as indicated in rows 10-13.

\subsection{Comparison against global motion SoA models}

Here we are testing the generalization of \model{} to predict global motion without a given scene. The task aims to be comparable to other scene-free methods.
The original version of \model{} is evaluated under the assumption of a ground surface beneath the individual. This assumption is implemented by converting the floor into a scene representation as a 3D point cloud. Based on this information, the model estimates future contact points. It is worth mentioning that unlike competing techniques such as \cite{Adeli2020SociallyAC, Wang2021MultiPerson3M}, our model does not include multi-person joint forecasting or consider social relationships among individuals.

\paragraph{Dataset.} We evaluate the performance of our model on additional datasets such as CMU-Mocap~\cite{CMU}, which is widely used for absolute pose forecasting. The CMU dataset is captured at a rate of 30 frames per second (fps) using a marker system. Each sequence in the dataset consists of three individuals randomly selected from different scenes and merged together~\cite{Wang2021MultiPerson3M}.

\paragraph{Comparison with state-of-the-art.} Our model is compared to SoA approaches, among which are HRI~\cite{mao2020history}, SocialPool~\cite{Adeli2020SociallyAC}, and MR-Trans~\cite{Wang2021MultiPerson3M}. HRI utilizes a motion attention mechanism to encode motion in both spatial coordinates and frequencies. SocialPool, on the other hand, is an RNN-based model that employs multiple GRU modules independently for each person in the scene, followed by a social module that considers the features of all individuals in the scene. MR-Trans, currently considered the SoA model, is a transformer-based approach that employs a discriminator to determine the suitability of pose and motion. Lastly, we also adapt Mao \etal~\cite{mao2022contact} to the additional dataset \textit{as is}.\\
The proposed approach performs similarly to the current best technique~\cite{Wang2021MultiPerson3M} in terms of overall error. However, when predicting future trajectories on the most challenging longer-term horizon, \model{} slightly underperforms compared to MR-Trans. On pose error, \model{} outperforms the previous SoA model~\cite{Wang2021MultiPerson3M}  by 33.5\%, and by 8.8\% with respect to~\cite{mao2022contact} on the most challenging longer-term horizon.

\begin{table}[!htp]
\centering

\scriptsize
\begin{tabular}{l|ccc|ccc|ccc}
\hline
\textbf{} &\multicolumn{3}{c|}{\textbf{Path Error}} &\multicolumn{3}{c|}{\textbf{Pose Error}}&\multicolumn{3}{c}{\textbf{Global Error}} \\
\textbf{Models} &\textbf{1s} &\textbf{2s} &\textbf{3s} &\textbf{1s} &\textbf{2s} &\textbf{3s}&\textbf{1s} &\textbf{2s} &\textbf{3s} \\
\hline
LTD~\cite{mao2019learning} & 0.97& 1.73& 2.62& 0.98& 1.21& 1.37& 1.37& 2.19& 3.26\\
HRI~\cite{mao2020history}& 0.96 & 2.06 & 3.11 & 1.05 & 1.37 & 1.58 & 1.49 & 2.60 & 3.07\\
SocialPool~\cite{Adeli2020SociallyAC}& 0.96 & 2.01 & 2.96 & 1.03 & 1.41 & 1.71 & 1.15 & 2.71 & 3.90\\
MR-Trans~\cite{Wang2021MultiPerson3M}& \textbf{0.60} & \textbf{1.12} & \textbf{1.71} & 0.79 & 1.05 & 1.22 & 0.96 & \textbf{1.57} & \textbf{2.18} \\
Mao \etal~\cite{mao2022contact}~\textit{w/ pred contact}& {0.78} & {2.19} & {3.99} & {0.59} & {0.93} & {0.95} & {1.01} & {2.47} & {4.16}\\
\model~\textit{w/ pred contact} & 0.71 & 1.43 & 2.02 & \textbf{0.57} & \textbf{0.76} & \textbf{0.87} & \textbf{0.95} & 1.70 & 2.29

\\ \hline

\end{tabular}
\caption{Path, pose and global error in meters on CMU-Mocap dataset.}
\label{tab: mocap}
\end{table}

\section{Conclusion}

This paper has addressed the prediction of global pose in a three-dimensional environment as the staged modelling of three core elements: the scene, the human trajectory, and the pose.
\model{} is the first scene-aware global forecasting model which splits trajectory and pose motion to match the coarse-to-fine nature of the task. In fact, the pose of a person is the result of its motion pathway and the scene, rather than the cause of it.

\model{} yields SoA performance on GTA-IM, the sole available for testing scene-aware global forecasting. \model{} also sets the SoA on the CMU-Mocap dataset, under the assumption that the scene consists solely of a flat ground surface, therefore generalizing the task of global forecasting, which earlier methods have addressed without consideration of the scene.

\paragraph{Acknowledgements} This work was supported by the MUR PNRR project FAIR (PE00000013).

\bibliography{egbib}

\newpage

\begin{center}
\textbf{\large Staged Contact-Aware Global Human Motion Forecasting    Supplementary Material}
\end{center}
\setcounter{equation}{0}
\setcounter{figure}{0}
\setcounter{table}{0}
\setcounter{page}{1}
\setcounter{section}{0}

\section{Implementation Details}
Our models are implemented in PyTorch and trained using the
ADAM~\cite{Kingma2014AdamAM} optimizer.
All of our stages: contact prediction, trajectory forecasting, and global pose forecasting, are trained for 50 epochs with learning rates of $5e-4$ and $1e-3$, respectively.
The training of each network takes about 12 hours on a 24GB NVIDIA RTX 3090 Ti GPU.
In line with \cite{mao2022contact}, we normalize data with a factor of $0.2$, using $20$ DCT coefficients and setting the contact threshold to $0.32$.
From each sequence, we sample $5000$ points in a range of 2.5 meters from the human pose's last observed root joint. 

\section{Remarks on the SoA on CMU-Mocap}

Sec.~4.3 and Table~4 in the main paper report published state-of-the-art (SoA) on the CMU-Mocap.
We report results from the current best performer MR-Trans~\cite{Wang2021MultiPerson3M}, from which we adopt the benchmark protocol and code.
These include \emph{Path Error}, \emph{Pose Error}, \emph{Global Error} (cf.\ Sec.~4.3).

We note that more recent unpublished work, SoMoFormer~\cite{vendrow2022somoformer}, reports to outperform MR-Trans on the task of \emph{Global Pose} but the error estimates do not match the MR-Trans paper~\cite{Wang2021MultiPerson3M}. 
In an effort to provide an overview of the SoA that is as complete as possible, we have re-run SoMoFormer~\cite{vendrow2022somoformer} within the same protocol as MR-Trans~\cite{Wang2021MultiPerson3M}, but we could not match their results.

Table~4 and Sec.~4.3 exclude SoMoFormer~\cite{vendrow2022somoformer} from the comparison since the work is not yet published, nor can we reproduce their results.
We commit to including the work in a later version of our manuscript, upon clarification of the mismatch.

\section{Qualitative Results}

\begin{figure}[h!t]
    \centering
    \includegraphics[width=0.8\textwidth]{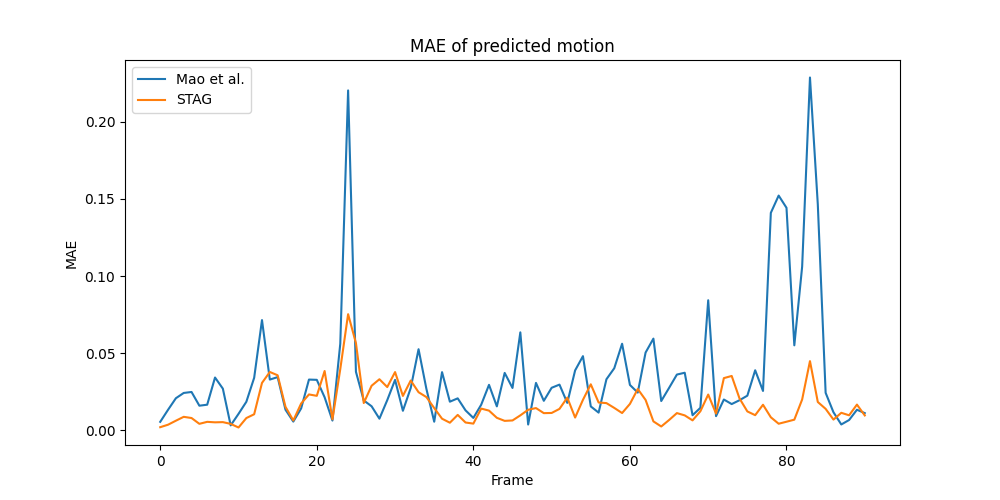}
    \caption{Comparison of the skeleton mean average error over root joints, at each timeframe, between \model{} and \cite{mao2022contact}.}
    \label{fig:frame_mae}
\end{figure}
\begin{figure}[!htp]
    \centering
    \begin{subfigure}[b]{0.45\textwidth}
        \includegraphics[width=\textwidth]{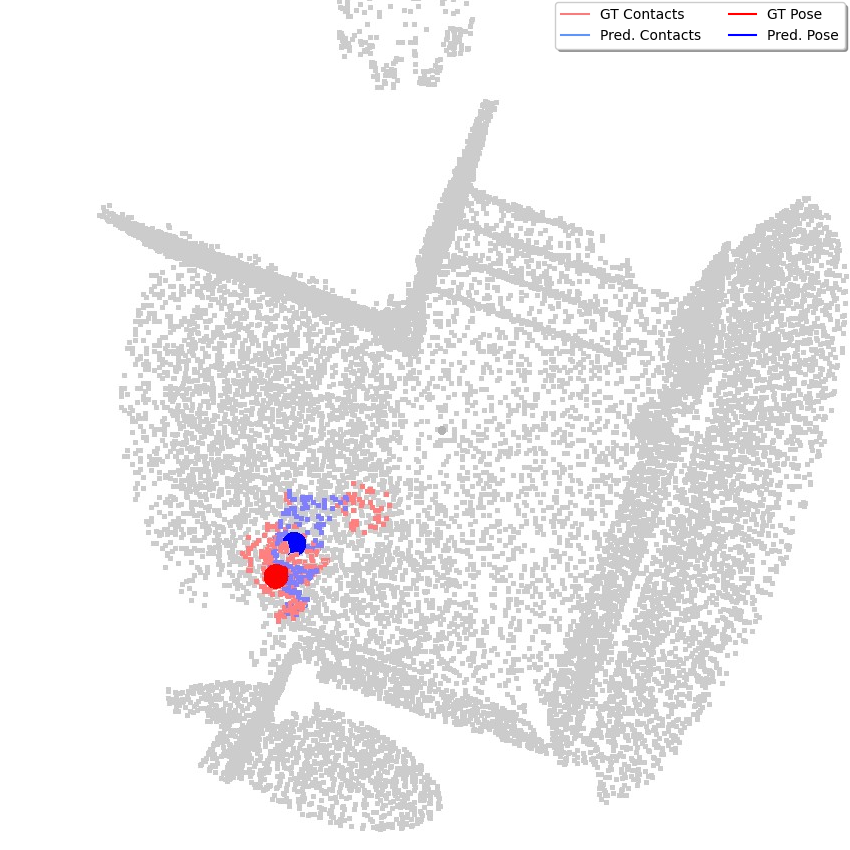}
        \caption{\model{}}
        \label{fig:top_ours}
    \end{subfigure}
    ~ 
    \begin{subfigure}[b]{0.45\textwidth}
        \includegraphics[width=\textwidth]{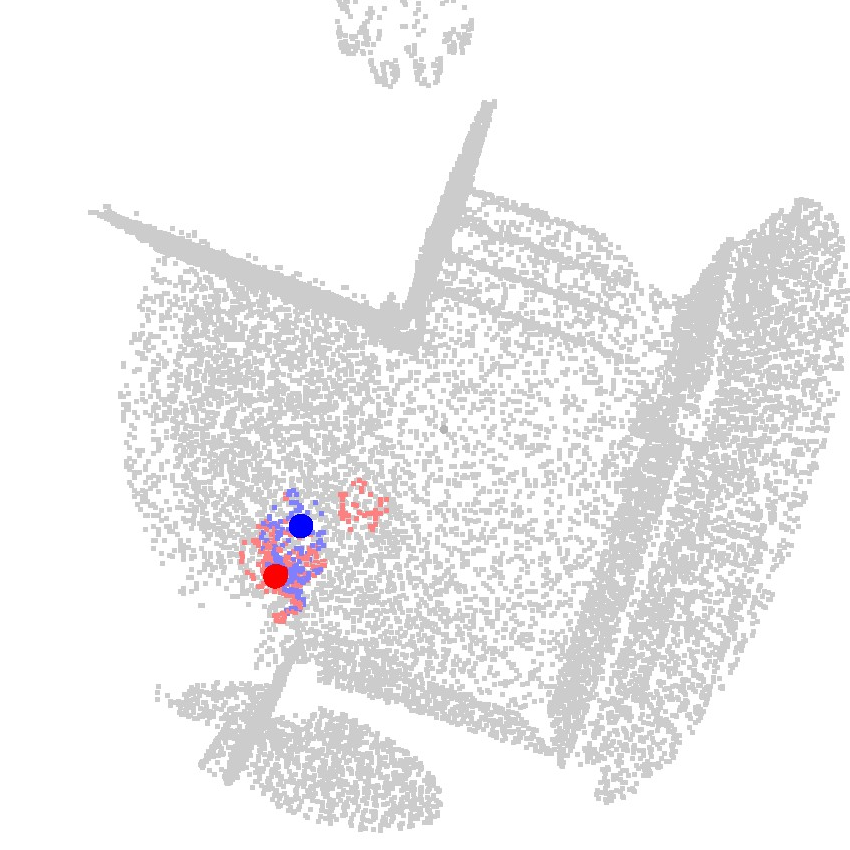}
        \caption{\model{} without temporal encoding.}
        \label{fig:top_ours:wo_te}
    \end{subfigure}
    \begin{subfigure}[b]{0.45\textwidth}
        \includegraphics[width=\textwidth]{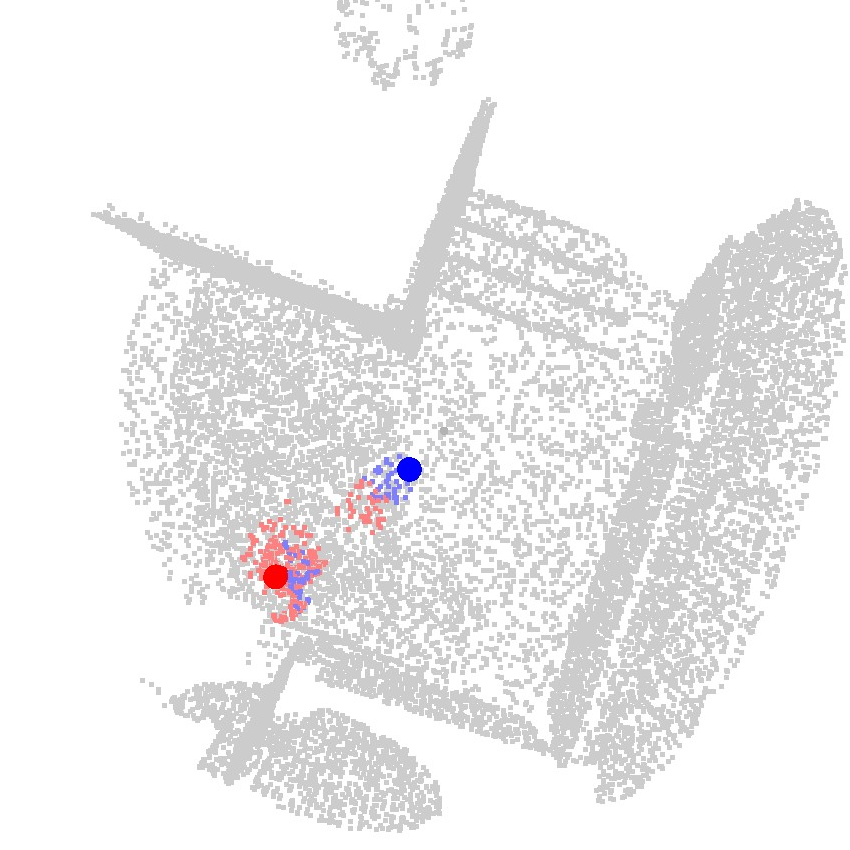}
        \caption{Mao \etal~\cite{mao2022contact}}
        \label{fig:top_mao}
    \end{subfigure}%
    \caption{Bird-eye view, which shows contact points and future root joint. 
    }
    \label{fig:top}
\end{figure}
\begin{figure}[!htp]
    \centering
    \begin{subfigure}[b]{0.45\textwidth}
        \includegraphics[width=\textwidth]{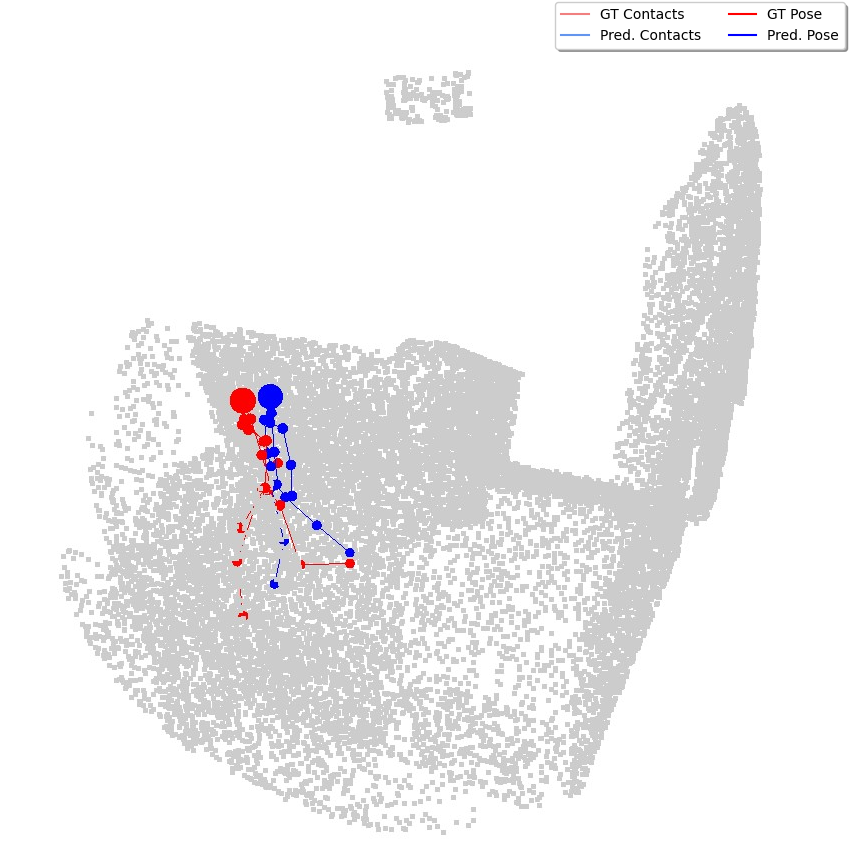}
        \caption{\model{}}
        \label{fig:sidewo_ours}
    \end{subfigure}%
    ~ 
    \begin{subfigure}[b]{0.45\textwidth}
        \includegraphics[width=\textwidth]{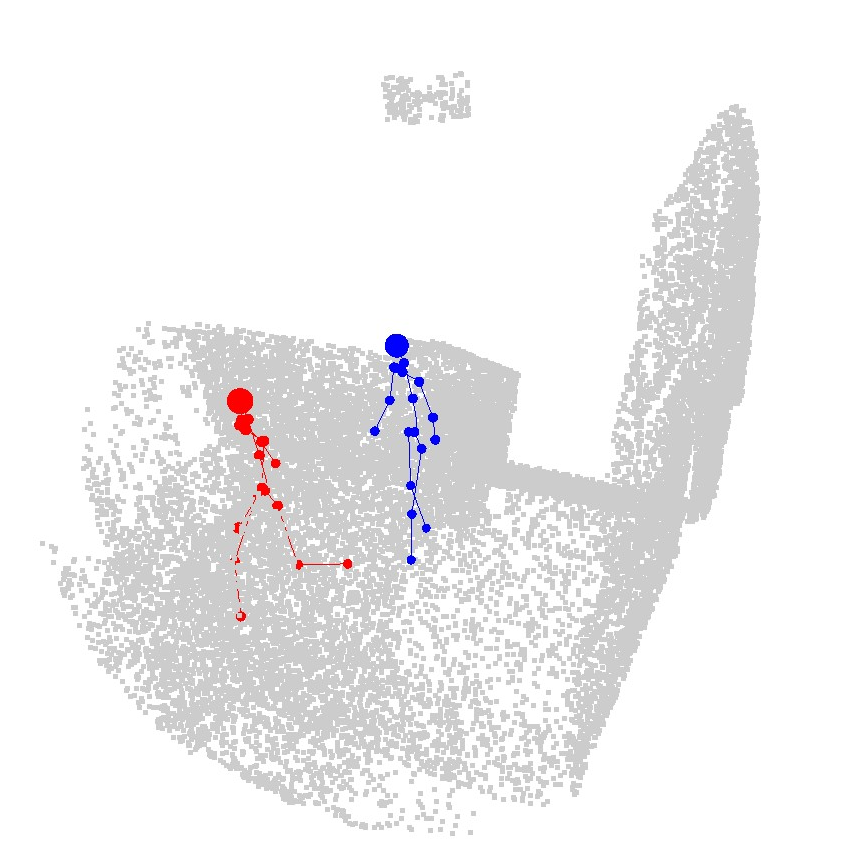}
        \caption{Mao \etal~\cite{mao2022contact}}
        \label{fig:sidewo_mao}
    \end{subfigure}
    \caption{Side view which shows the future global pose. 
    }\label{fig:sidewo}
\end{figure}
\begin{figure}[!htp]
    \centering
    \begin{subfigure}[b]{0.45\textwidth}
        \includegraphics[width=\textwidth]{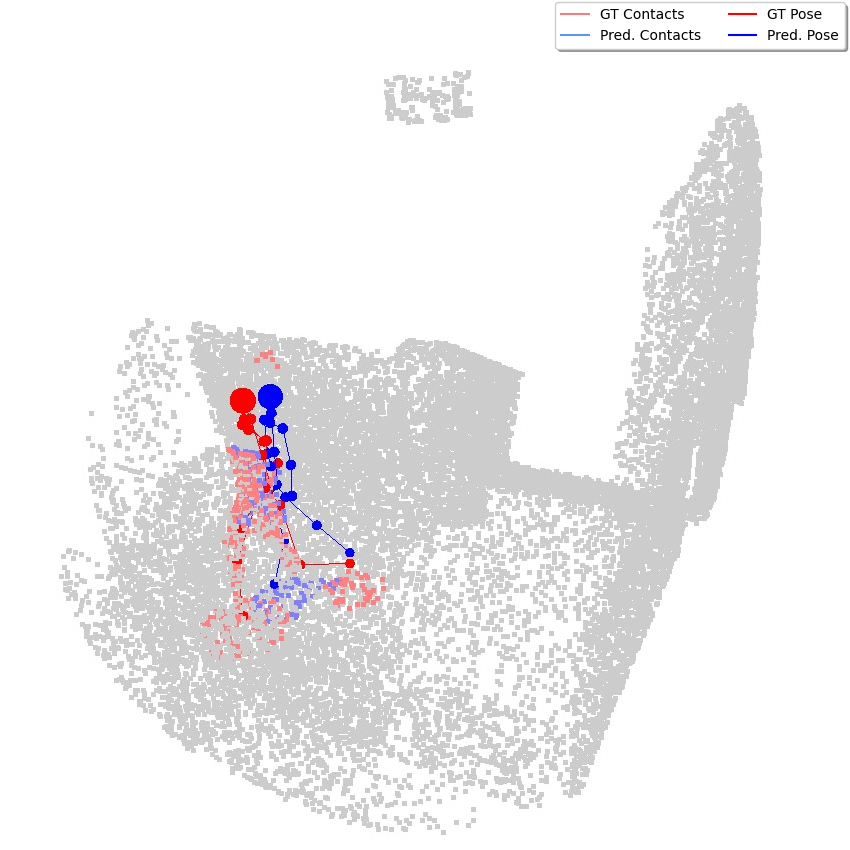}
        \caption{\model{}}
        \label{fig:side2_ours}
    \end{subfigure}%
    ~ 
    \begin{subfigure}[b]{0.45\textwidth}
        \includegraphics[width=\textwidth]{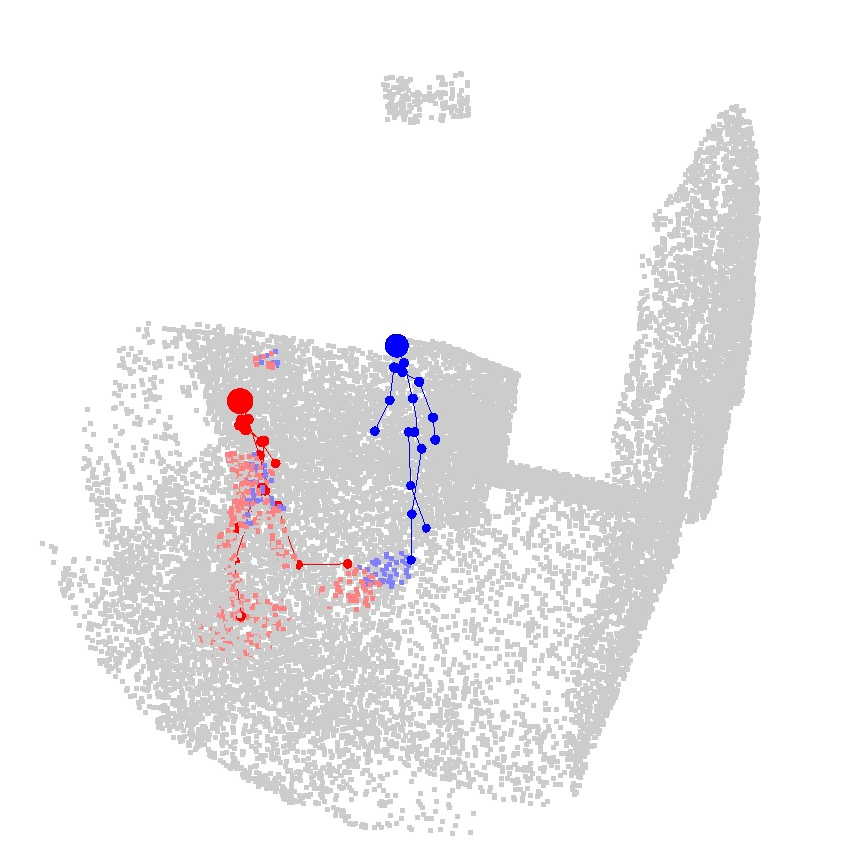}
        \caption{Mao \etal~\cite{mao2022contact}}
        \label{fig:side2_mao}
    \end{subfigure}
    \caption{Side view which shows contact points and future global pose.}\label{fig:side2}
\end{figure}
\begin{figure}[!htp]
    \centering
    \begin{subfigure}[b]{0.45\textwidth}
        \includegraphics[width=\textwidth]{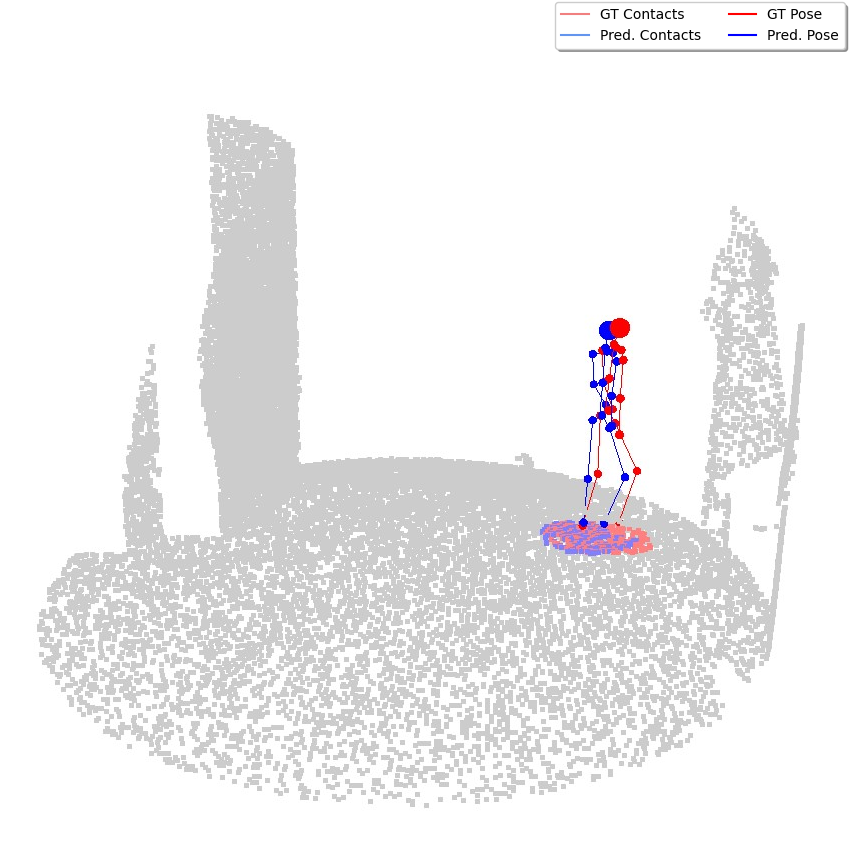}
        \caption{\model{}}
        \label{fig:side_ours}
    \end{subfigure}
    ~
    \begin{subfigure}[b]{0.45\textwidth}
        \includegraphics[width=\textwidth]{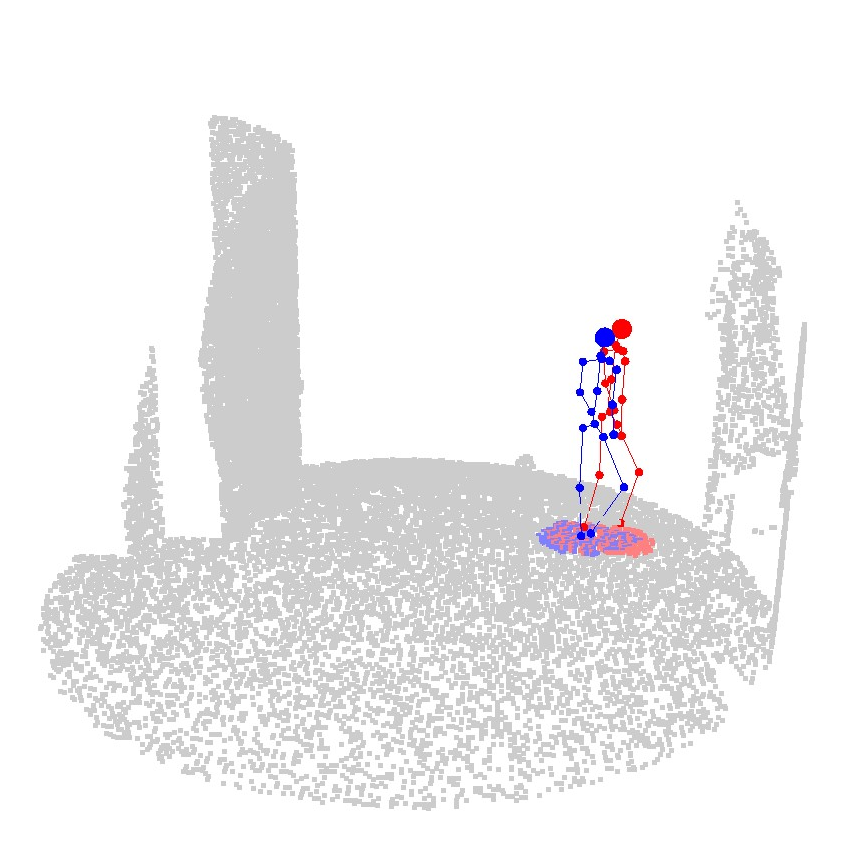}
        \caption{Mao \etal~\cite{mao2022contact}}
        \label{fig:side_mao}
    \end{subfigure}%
    \caption{Side view which shows contact points and future global pose.}\label{fig:side}
\end{figure}

We present visualizations obtained from the videos of both our model and the model proposed by Mao \etal~\cite{mao2022contact} in Figures (Fig. \ref{fig:top}-\ref{fig:side}).
These figures display the results at the final frame, precisely at 2 seconds, deliberately emphasizing the challenging nature of the prediction task.
In Fig. \ref{fig:frame_mae} we can see that our model generally performs better than \cite{mao2022contact} at the majority of frames.
The samples were randomly selected and it is worth mentioning that we noticed a posteriori how they fall in the top $34\%$ predictions.
One common issue of our predictions is the tendency of aggregating contact points (see Figs. \ref{fig:top_ours}, \ref{fig:top_ours:wo_te}), without discriminating between left and right foot; while \cite{mao2022contact} does not experience that issue (Fig. \ref{fig:top_mao}).
Yet, it seems a posteriori that not having disjoint contact point predictions guides better the human pose forecasting model, as it conditions on less noisy data.
In these visualizations, the ground truths are represented in shades of red, whereas the predictions are represented in shades of blue.
This applies to both the contact point and global human poses. 
Additionally, the scene is portrayed as a 3D Point Cloud.

Figures \ref{fig:top_ours}-\ref{fig:top_mao} provide a bird's-eye view of the scene, highlighting the predicted trajectory and contact points.
The model proposed by \cite{mao2022contact} tends to perceive each foot's contact points individually, while our model, referred to as \model{} (Figure \ref{fig:top_ours}), also considers the area between the feet. 
Consequently, our model demonstrates superior trajectory prediction, avoiding confusion caused by individual contact areas. 
It is worth mentioning that without our time-to-go temporal encoding, both the quantitative results (Table 3 of the main paper) and the qualitative results (Figure \ref{fig:top_ours:wo_te}) exhibit poorer performance. 
This is because the model lacks an understanding of the remaining time required to reach the endpoint.

The side view is beneficial for a more convenient display of global human pose forecasting. 
In Figures \ref{fig:sidewo}-\ref{fig:side}, we can observe the improved prediction of our model's human pose, both with and without predicted contact points. 
Figures \ref{fig:sidewo_ours} and \ref{fig:sidewo_mao} clearly illustrate how the second stage of \model{} contributes to better root positions.

Furthermore, even when considering similar contact points and root positions, our model consistently predicts more coherent human poses, as depicted in Figures \ref{fig:side_mao} and \ref{fig:side_ours}.

\end{document}